\documentclass{article}

 \usepackage[preprint]{neurips_2025}

\usepackage[utf8]{inputenc} 
\usepackage[T1]{fontenc}    
\usepackage{hyperref}       
\usepackage{url}            
\usepackage{booktabs}       
\usepackage{amsfonts}       
\usepackage{nicefrac}       
\usepackage{microtype}      
\usepackage{xcolor}         
\usepackage{graphicx}
\title{Solving Context Window Overflow in AI Agents}

\author{%
 Anton Bulle Labate \\
 IBM Research, Brazil\\
 \texttt{anton.labate1@ibm.com} \\
  \And
  Valesca Moura de Sousa \\
 IBM Research, Brazil\\
  \texttt{valesca.sousa@ibm.com} \\
  \And
  Sandro Rama Fiorini \\
  IBM Research, Brazil\\
  \texttt{srfiorini@ibm.com} \\
  \And
  Leonardo Guerreiro Azevedo \\
   IBM Research, Brazil\\
  \texttt{lga@br.ibm.com} \\
  \And
  Raphael Melo Thiago \\
  IBM Research, Brazil\\
  \texttt{raphaelt@br.ibm.com} \\
   \And
  Viviane Torres da Silva \\
  IBM Research, Brazil\\
  \texttt{vivianet@br.ibm.com} \\
}

\begin{document}

\maketitle

\begin{abstract}
   Large Language Models (LLMs) have become increasingly capable of interacting with external tools, granting access to specialized knowledge beyond their training data -- critical in dynamic, knowledge-intensive domains such as Chemistry and Materials Science. However, large tool outputs can overflow the LLMs' context window, preventing task completion. Existing solutions such as truncation or summarization fail to preserve complete outputs, making them unsuitable for workflows requiring the full data. This work introduces a method that enables LLMs to process and utilize tool responses of arbitrary length without loss of information. By shifting the model’s interaction from raw data to memory pointers, the method preserves tool functionality, allows seamless integration into agentic workflows, and reduces token usage and execution time. The proposed method is validated on a real-world Materials Science application that cannot be executed with conventional workflows, and its effectiveness is demonstrated via a comparative analysis where both methods succeed. In this experiment, the proposed approach consumed approximately seven times fewer tokens than the traditional workflow.
\end{abstract}

\section{Introduction}
Recent advances in Large Language Models (LLMs) have significantly enhanced their capabilities in tool usage.
This ability is particularly relevant, as it enables the LLMs to be applied/operate effectively in domains that were either underrepresented or entirely absent during their pretraining phase.
By leveraging domain-specific knowledge through tool outputs, LLMs can overcome limitations in their original training data.
When prompted with a request for which they require auxiliary knowledge to answer, the LLMs just need to identify the tool that best suits them in the case and call it with the appropriate arguments. The tool result is subsequently incorporated into the LLM's context window (i.e., the information that is in fact considered by the LLM when generating a response), thereby enabling it to reason over this augmented information -- either for directly answering the user request, or use it for further processing (by another tool, for instance). 
This capability unlocks their potential in dynamic and knowledge-intensive fields such as chemistry research and laboratory assistants\cite{boiko2023autonomous, chemcrow}.

Nonetheless, when the tool outputs a large response, the authors of LongFuncEval\cite{kate2025longfuncevalmeasuringeffectivenesslong} show that frontier models' performance deteriorates drastically (from 7\% to about 91\%). In more extreme scenarios, the length of certain tool-generated outputs may even surpass the model’s context window, thereby limiting their practical applicability in domains characterized by such demands. This constraint is particularly salient in disciplines such as Materials Science and Chemistry, where voluminous outputs are frequently encountered.

As solutions to this context window limitation within the scope of tool calling without requiring architectural changes, works have explored strategies such as summarizing tool descriptions \cite{xu-etal-2024-concise} or selectively including tools in the context \cite{zhang2025ecoact, qin2024toolllm, lumer2024toolshed}, tackling the issue of a large tool catalog. 

Other methods focus on parsing and truncating tool outputs to manage excessively large responses \cite{song2023restgpt}. However, none of these approaches is able to deal with large, indivisible outputs. Since all of them present a loss of some sort of the tool response--such as truncating or summarizing it--they are unable to preserve and utilize the tool output in its original, complete form as input for subsequent processing steps, thereby hindering their application to workflows dependent on such outputs for further processing.

In this work, we introduce a method designed to enable LLMs to process tool responses of arbitrary length entirely — including those that far exceed the model’s context window — and to subsequently use them as input for other tools or as output to the user, without any information loss. The agent accomplishes this by operating on pointers to large data values rather than the raw data itself.
This approach preserves both the functional behavior and input expectations of the original tools, ensuring full transparency to the agent and seamless integration within any tool or agentic framework. Additionally, by manipulating pointers instead of raw data, the proposed method reduces token generation compared to conventional workflows, resulting in lower computational costs and faster decoding times. Crucially, no architectural modifications or changes to the underlying tools are necessary.

With our method, we extend the applicability of agent systems to domains and use cases in which large tool responses are routine, such as Materials Science and Chemistry, without loss of information. To demonstrate this, we apply the approach in a real-world Materials Science application, which could not be accomplished by a conventional agentic workflow.

\section{Handling large tool responses}

In many real-world domains, it is common for tools to receive and generate extensive data. In the context of Materials Discovery, for instance,
an agent might be tasked to generate a molecule structure and reason about its properties--potentially using these structures as inputs to other functions to filter out undesirable candidates. 
Applications like this cannot be solved using a conventional agentic workflow that presents these tool outputs directly to the language model in its context, as it would overflow its context window. Proposed techniques of extending the context window of LLMs without architectural changes also cannot solve the issue, as they circumvent this problem by treating a large input as divisible, and therefore do not present it in its entirety to the model.

\begin{figure}[h]
    \centering
    \includegraphics[width=1.0\linewidth]{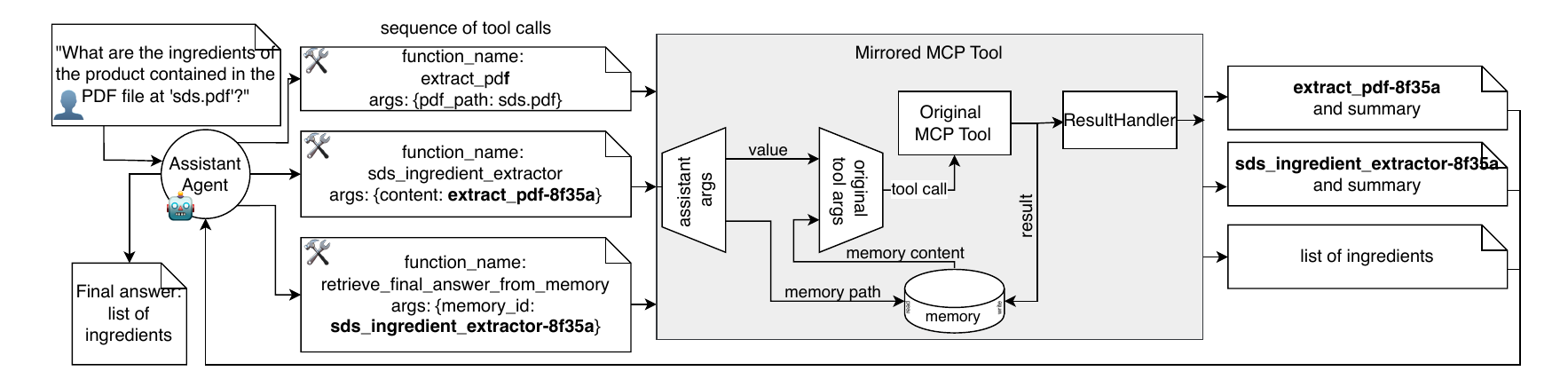}
    \caption{Overview of the approach: general architecture of the proposed method and workflow illustration of a real-world use case.}
    \label{fig:diagram}
\end{figure}

To overcome this limitation, we store large data outside the LLM's context window, and allow the model to interact with it using only short identifiers (i.e., pointers). Each tool in the agent’s toolkit--whether provided by an MCP server or defined locally--is mirrored with a wrapper that enables it to handle these pointers. A mirrored tool comprises three components: (1) a module that inspects the input to determine whether it is a pointer or a raw value, (2) the original tool with its unmodified behavior, and (3) a post-processing module that stores large outputs in memory and returns access instructions if necessary. A high-level overview of the solution is illustrated by the \textit{MCP Mirrored Tool} component in Figure~\ref{fig:diagram}.

During agent execution, a runtime memory is maintained to temporarily store large values that exceed the model’s context window. Each stored value is assigned a unique path, constructed from the name of the tool that produced the output and a corresponding identifier. In cases in which the tool returns a dictionary containing multiple key-value pairs, each pair is stored individually under the same base path, extended with the respective key. This hierarchical naming scheme ensures accurate and efficient retrieval of stored data during subsequent interactions.

Upon receiving input, a mirrored tool first determines whether the input corresponds to a memory path and resolves any pointers to their associated values. Once all arguments have been associated to their actual content, they are forwarded to the original tool for execution. The resulting output is then stored in the runtime memory under a path that adheres to the established naming convention.
To present final results to the user, a dedicated tool retrieves values from memory when explicitly needed.

Using this approach, the LLM can invoke any tool it chooses, using any arguments it specifies, without concern for exceeding the context window. 
The model interacts with data at a higher abstraction level--manipulating references to data rather than the data itself--preserving functionality while enabling scalability.

\section{Experiments}

To illustrate the practical value of the method, a representative use case from the Materials Science domain was selected: given a molecule, the user requests a set of molecules with similar electronic grid structures. This scenario exemplifies the type of problem the approach is designed to address. Importantly, the method must also remain effective when tool outputs are small enough to fit within the model’s context window. To validate this, a second experiment was conducted in which the agent processes a Safety Data Sheet (SDS) provided in PDF format and accurately identifies the ingredients listed within the document. 

Both experiments were executed using an agent powered by Llama 4 Maverick 17B-128E-Instruct\footnote{https://huggingface.co/meta-llama/Llama-4-Maverick-17B-128E-Instruct} and the BeeAI agentic framework,\footnote{https://github.com/i-am-bee/beeai-framework} guided by a simple system prompt defining its role as a helpful scientific assistant. The agent followed the ReAct pattern and was equipped with the relevant MCP tools for each task.

\subsection{Retrieval of molecules with similar electronic grid structures} \label{exp:molecules}

In this experiment, the agent is equipped with two tools: one that generates the electronic grid structure of a molecule, and another that retrieves molecules with similar electronic grids. The first tool takes a molecule’s SMILES string as input and outputs a 3D matrix of dimensions (128, 128, 128), totaling 2,097,152 float32 elements. This output far exceeds the context window of any widely used LLM. The second tool accepts such an electronic grid as input and returns a list of the top-k most similar molecules including their similarity scores and SMILES strings. 

 Although the output of the first tool serves as input to the second, we maintain them as separate tools to reflect realistic agentic settings, in which outputs may be reused across multiple tools. Presenting only the relevant tools to the agent simplifies the task, but our goal is not to evaluate tool selection or planning capabilities--instead, we focus on how the model \emph{handles} large tool responses.

In this experiment, the user asks the agent to identify the molecules whose electronic grids are similar to that of the query molecule, specified via the provided SMILES string. The agent should first call the grid-generation tool with the user-defined SMILES as input, then use the resulting electronic grid it outputs  as input to the similarity-retrieval tool.

Under the conventional workflow, the agent successfully calls the first tool with the appropriate argument. Nevertheless, as the tool response exceeds (by far) the model's maximum input length, an error occurs due to the message's body size exceeding its given limit, failing at this stage.

In contrast, with the proposed method, the agent is able to complete the entire pipeline. After invoking the grid generation tool, the agent receives a memory path indicating where the output is stored and how it can be accessed. This reference allows the agent to interact with the electronic grid as if it were directly available within its context window. Instead of passing the raw grid data to subsequent tools, the agent passes the memory path, enabling seamless integration into the workflow despite the data’s size. A complete execution trace of this task using the proposed method is provided in the Appendix~\ref{ap:molecules}. 

The complete execution using the proposed method took an average of 33.87 seconds and 1,234.04 tokens, based on 50 runs with identical user and system prompts. Because with the conventional approach the agent was unable to complete the workflow, we couldn't measure execution time in that case. However, we can estimate the token usage up to the point where the context overflow occurred. By substituting the memory path returned in the successful executions with the actual tool output, we approximate the number of tokens that would have been consumed using the conventional approach up to that stage in the workflow. Applying this substitution across all 50 runs yields an average token consumption of 
20,822,181.34 tokens, highlighting the infeasibility of the conventional approach for handling large tool outputs.

\subsection{Extracting ingredients from a safety data sheet}

To perform the extraction of ingredient information from Safety Data Sheets (SDS), the agent has access to two complementary tools: (i) a PDF extractor, which receives the path to a PDF file, extracts its content, and returns it as a text string; and (ii) an SDS ingredient extractor, which processes the textual content of the SDS, identifies and returns relevant information regarding the ingredients.

Figure~\ref{fig:diagram} illustrates the workflow execution following our proposed approach. The user provides a prompt requesting the ingredients contained in a PDF file (e.g., ``What are the ingredients of the product contained in the PDF file at `sds.pdf'?''). The agent begins by selecting the PDF extraction tool and providing `sds.pdf' as input, which returns a memory path referencing the extracted content. Using this memory path, the agent then runs the SDS ingredient extractor. Within the Mirrored MCP tool, the memory path retrieves the stored content and executes the tool with its original parameters, producing another memory path pointing to the extracted ingredients. Finally, the agent calls the tool to retrieve the final answer and presents the result to the user. A detailed execution trace of this workflow, illustrating the proposed method, is included in the Appendix~\ref{ap:extraction}.

The experiment involved executing the workflows 50 times to collect metrics on execution time and token usage, both using the proposed solution and the traditional approach (i.e., without the proposed solution). In the traditional approach, an average of 6,411.36 tokens were used, with an execution time of 43.05 seconds. In contrast, executing the workflow with the proposed solution required an average of 841.66 tokens and 11.05 seconds. These results demonstrate that the solution, in addition to enabling tool calls that would exceed an LLM agent’s context window as described in Experiment~\ref{exp:molecules}, does not interfere with workflows where the LLM’s context window would otherwise be sufficient. On the contrary, it serves as an effective mechanism to significantly reduce token usage --approximately sevenfold in this use case -- thereby accelerating workflow execution, since only the most relevant information from tool outputs is passed to the LLM. Furthermore, the solution provides cost savings in scenarios where LLMs are deployed in cloud services, as the cost in such contexts is directly related to the number of tokens processed.

\section{Conclusion}

In applications such as Accelerated Discovery and Materials Design, tool outputs can exceed the context window of LLMs, causing overflow and limiting the applicability of agentic workflows. This work introduces a method that handles large tool responses by shifting the model’s tools domain from raw data to pointers referencing memory-stored results, preserving tool functionality and enabling scalability to arbitrarily large inputs. Consequently, agentic systems can tackle previously infeasible use cases.
Furthermore, the proposed method offers additional benefits beyond handling large data. As demonstrated, it reduces token consumption-- which directly translates into cost savings when using cloud-based LLM services -- and improves execution speed even in workflows where tool outputs fit within the model's context window. This makes the approach broadly applicable and efficient across a wide range of scenarios.
For future work, the agent should be equipped with a mechanism to transform the data schema of a memory element. For those cases, we probably need to devise ways for the agent to selectively access verbatim parts of a memory element within context restrictions.

\bibliographystyle{unsrt}
\bibliography{references}  

@article{chemcrow,
  title={Augmenting large language models with chemistry tools},
  author={M. Bran, Andres and Cox, Sam and Schilter, Oliver and Baldassari, Carlo and White, Andrew D and Schwaller, Philippe},
  journal={Nature Machine Intelligence},
  volume={6},
  number={5},
  pages={525--535},
  year={2024},
  publisher={Nature Publishing Group UK London}
}

@article{boiko2023autonomous,
  title={Autonomous chemical research with large language models},
  author={Boiko, Daniil A and MacKnight, Robert and Kline, Ben and Gomes, Gabe},
  journal={Nature},
  volume={624},
  number={7992},
  pages={570--578},
  year={2023},
  publisher={Nature Publishing Group UK London}
}

@inproceedings{xu-etal-2024-concise,
    title = "Concise and Precise Context Compression for Tool-Using Language Models",
    author = "Xu, Yang  and
      Feng, Yunlong  and
      Mu, Honglin  and
      Hou, Yutai  and
      Li, Yitong  and
      Wang, Xinghao  and
      Zhong, Wanjun  and
      Li, Zhongyang  and
      Tu, Dandan  and
      Zhu, Qingfu  and
      Zhang, Min  and
      Che, Wanxiang",
    editor = "Ku, Lun-Wei  and
      Martins, Andre  and
      Srikumar, Vivek",
    booktitle = "Findings of the Association for Computational Linguistics: ACL 2024",
    month = aug,
    year = "2024",
    address = "Bangkok, Thailand",
    publisher = "Association for Computational Linguistics",
    url = "https://aclanthology.org/2024.findings-acl.974/",
    doi = "10.18653/v1/2024.findings-acl.974",
    pages = "16430--16441"
}

@article{lumer2024toolshed,
  title={Toolshed: Scale tool-equipped agents with advanced rag-tool fusion and tool knowledge bases},
  author={Lumer, Elias and Subbiah, Vamse Kumar and Burke, James A and Basavaraju, Pradeep Honaganahalli and Huber, Austin},
  journal={arXiv preprint arXiv:2410.14594},
  year={2024}
}

@inproceedings{
qin2024toolllm,
title={Tool{LLM}: Facilitating Large Language Models to Master 16000+ Real-world {API}s},
author={Yujia Qin and Shihao Liang and Yining Ye and Kunlun Zhu and Lan Yan and Yaxi Lu and Yankai Lin and Xin Cong and Xiangru Tang and Bill Qian and Sihan Zhao and Lauren Hong and Runchu Tian and Ruobing Xie and Jie Zhou and Mark Gerstein and dahai li and Zhiyuan Liu and Maosong Sun},
booktitle={The Twelfth International Conference on Learning Representations},
year={2024},
url={https://openreview.net/forum?id=dHng2O0Jjr}
}

@article{song2023restgpt,
  title={Restgpt: Connecting large language models with real-world restful apis},
  author={Song, Yifan and Xiong, Weimin and Zhu, Dawei and Wu, Wenhao and Qian, Han and Song, Mingbo and Huang, Hailiang and Li, Cheng and Wang, Ke and Yao, Rong and others},
  journal={arXiv preprint arXiv:2306.06624},
  year={2023}
}

@misc{kate2025longfuncevalmeasuringeffectivenesslong,
      title={LongFuncEval: Measuring the effectiveness of long context models for function calling}, 
      author={Kiran Kate and Tejaswini Pedapati and Kinjal Basu and Yara Rizk and Vijil Chenthamarakshan and Subhajit Chaudhury and Mayank Agarwal and Ibrahim Abdelaziz},
      year={2025},
      eprint={2505.10570},
      archivePrefix={arXiv},
      primaryClass={cs.SE},
      url={https://arxiv.org/abs/2505.10570}, 
}

@inproceedings{
zhang2025ecoact,
title={EcoAct: Economic Agent Determines When to Register What Action},
author={Shaokun Zhang and Jieyu Zhang and Dujian Ding and Jiale Liu and Mirian Del Carmen Hipolito Garcia and Ankur Mallick and Daniel Madrigal and Menglin Xia and Victor R{\"u}hle and Qingyun Wu and Chi Wang},
booktitle={Workshop on Reasoning and Planning for Large Language Models},
year={2025},
url={https://openreview.net/forum?id=IHgVuYwhnz}
}

\appendix
\section{Workflow execution of the retrieval of molecules with similar electronic grid structures experiment} \label{ap:molecules}

\begin{verbatim}
RequirementAgent[RequirementAgent][start]:
{"input": {"prompt": "For the molecule in the grid, with SMILES 
OC12COC3=NCC1C23, what are the most similar molecules to it?"}}

-->  ThinkTool[think][start]: 
{"input": {"input": {"next_step": ["generate_molecule_grid_mirrored"], 
"thoughts": "To find the most similar molecules to the given molecule 
with SMILES OC12COC3=NCC1C23, I need to first generate its electronic 
grid using the `generate_molecule_grid_mirrored' tool"}}}

<--  WatsonxChatModel[finish]: 
{"messages": [{"role": "assistant", "content": [{"type": "tool-call", 
"id": "chatcmpl-tool-773240223f9b470ca20d92551a96bfc7", 
"tool_name": "generate_molecule_grid_mirrored", 
"args": "{\"molecule_description\": \"OC12COC3=NCC1C23\"}"}]}], 
"usage": {"prompt_tokens": 1576, "completion_tokens": 18, 
"total_tokens": 1594}, "finish_reason": "stop"}

-->  FunctionTool[generate_molecule_grid_mirrored][start]: 
{"input": {"input": {"molecule_description": "OC12COC3=NCC1C23"}}}
Processing request of type CallToolRequest

<--  FunctionTool[generate_molecule_grid_mirrored][finish]: 
"The result of the function generate_molecule_grid with the input value 
stored at {\"molecule_description\": \"OC12COC3=NCC1C23\"} is currently 
stored at generate_molecule_grid-fcb87ffa-31b7-41b0-bf90-76d0c87000f5.
When you need to access it, pass as argument for the tool its path: 
generate_molecule_grid-fcb87ffa-31b7-41b0-bf90-76d0c87000f5. 
The result stored is a dict with the keys raw_grid and shape. 
When you want to access only a particular value stored under these keys, 
use as memory path 
generate_molecule_grid-fcb87ffa-31b7-41b0-bf90-76d0c87000f5/raw_grid
or generate_molecule_grid-fcb87ffa-31b7-41b0-bf90-76d0c87000f5/shape, 
depending on which value you want to use."

<--  WatsonxChatModel[finish]: 
{"messages": [{"role": "assistant", "content": [{"type": "tool-call", 
"id": "chatcmpl-tool-6d3bfbf8faf34190827462d105c99611", 
"tool_name": "think", "args": 
"{\"next_step\": [\"retrieve_similar_molecules_mirrored\"], 
\"thoughts\": \"Now that I have the electronic grid for the molecule with 
SMILES OC12COC3=NCC1C23, I can use the 
'retrieve_similar_molecules_mirrored' tool to find the most 
similar molecules to it.\"}"}]}], 
"usage": {"prompt_tokens": 2569, "completion_tokens": 78, 
"total_tokens": 2647}, "finish_reason": "tool_calls"}

--> ThinkTool[think][start]: 
{"input": {"input": {"next_step": 
["retrieve_similar_molecules_mirrored"], 
"thoughts": "Now that I have the electronic grid for the molecule with 
SMILES OC12COC3=NCC1C23, I can use the 
'retrieve_similar_molecules_mirrored' tool to find the most 
similar molecules to it."}}}

<-- WatsonxChatModel[finish]: 
{"messages": [{"role": "assistant", "content": [{"type": "tool-call", 
"id": "chatcmpl-tool-e05460e44eba4efb8fb92d6220a25daf", 
"tool_name": "retrieve_similar_molecules_mirrored", 
"args": "{\"raw_grid\": 
\"generate_molecule_grid-fcb87ffa-31b7-41b0-bf90-76d0c87000f5/raw_grid\"
}"}]}], 
"usage": {"prompt_tokens": 2464, "completion_tokens": 54,
"total_tokens": 2518}, "finish_reason": "tool_calls"}

-->  FunctionTool[retrieve_similar_molecules_mirrored][start]:
{"input": {"input": {"raw_grid": 
"generate_molecule_grid-fcb87ffa-31b7-41b0-bf90-76d0c87000f5/raw_grid"}
}}

<--  FunctionTool[retrieve_similar_molecules_mirrored][finish]: 
"The result of the function retrieve_similar_molecules with the input
 value 
stored at {\"raw_grid\": 
\"generate_molecule_grid-fcb87ffa-31b7-41b0-bf90-76d0c87000f5/raw_grid\
"} is currently stored at 
retrieve_similar_molecules-30daddd0-d4a1-4689-bc78-32eb93b16252."

<-- WatsonxChatModel[finish]: 
{"messages": [{"role": "assistant", "content": [{"type": "tool-call", 
"id": "chatcmpl-tool-4b998bb8096d453493936673c35685b9", 
"tool_name": "retrieve_final_answer_from_memory", "args": 
"{\"memory_path\":
\"retrieve_similar_molecules-30daddd0-d4a1-4689-bc78-32eb93b16252\"}"}]
}], "usage": {"prompt_tokens": 2857, "completion_tokens": 49, 
"total_tokens": 2906}, "finish_reason": "tool_calls"}

<-- FunctionTool[retrieve_final_answer_from_memory][finish]: 
"Top-K similar samples:
1. SMILES: C1CC1CC1CC1 | Score: 0.4499
2. SMILES: CC(=O)C1C2CCC21 | Score: 0.4251
3. SMILES: CC1C(O)C1(C)C#N | Score: 0.4187
4. SMILES: COC12CC(C)OC1C2 | Score: 0.4097
5. SMILES: O=CC1CCC1CCO | Score: 0.4093
6. SMILES: C#CCC(=O)C(C)C | Score: 0.4041
7. SMILES: CC(C)C12CCC1C2 | Score: 0.4014
8. SMILES: CC12COC(O1)C(O)C2 | Score: 0.3900
9. SMILES: CC12CC3(O)C(O)C1C23 | Score: 0.3840
10. SMILES: CC(C#N)NCC1CO1 | Score: 0.3734"

The most similar molecules to the molecule with SMILES OC12COC3=NCC1C23 
are:
1. SMILES: C1CC1CC1CC1 | Score: 0.4499
2. SMILES: CC(=O)C1C2CCC21 | Score: 0.4251
3. SMILES: CC1C(O)C1(C)C#N | Score: 0.4187
4. SMILES: COC12CC(C)OC1C2 | Score: 0.4097
5. SMILES: O=CC1CCC1CCO | Score: 0.4093
6. SMILES: C#CCC(=O)C(C)C | Score: 0.4041
7. SMILES: CC(C)C12CCC1C2 | Score: 0.4014
8. SMILES: CC12COC(O1)C(O)C2 | Score: 0.3900
9. SMILES: CC12CC3(O)C(O)C1C23 | Score: 0.3840
10. SMILES: CC(C#N)NCC1CO1 | Score: 0.3734
\end{verbatim}

\section{Workflow execution of the extracting ingredients from a safety data sheet experiment} \label{ap:extraction}

\begin{verbatim}
RequirementAgent[RequirementAgent][start]: 
{"input": {"prompt": "What are the ingredients of the product 
contained in the PDF file at 'sds.pdf'? Describe the name, chemical 
formula and CAS number of ingredients, if any."}}

<-- WatsonxChatModel[finish]:
{"messages": [{"role": "assistant", "content": [{"type": "tool-call",
"id": "chatcmpl-tool-a7e91ad199444813b752b064a913f743", 
"tool_name": "tika_mirrored", "args": "{\"pdf_path\": 
\"sds.pdf\"}"}]}], "usage": {"prompt_tokens": 1184, 
"completion_tokens": 37, "total_tokens": 1221}, 
"finish_reason": "stop"}

--> FunctionTool[tika_mirrored][start]:
{"input": {"input": {"pdf_path": "sds.pdf"}}}

<-- FunctionTool[tika_mirrored][finish]:
"The result of the function tika with the input value stored at 
{\"pdf_path\": \"sds.pdf\"} is currently stored at 
tika-d719493f-b573-4dc2-b15c-6d031f64b7af. When you need to access 
it, pass as argument for the tool its path: 
tika-d719493f-b573-4dc2-b15c-6d031f64b7af."

<-- WatsonxChatModel[finish]:
{"messages": [{"role": "assistant", "content": [{"type": "tool-call",
"id": "chatcmpl-tool-5b0e976157bf474bb18f8f0fe23fccfa", 
"tool_name": "extract_sds_mirrored", "args": "{\"content\": 
\"tika-d719493f-b573-4dc2-b15c-6d031f64b7af\"}"}]}], "usage": 
{"prompt_tokens": 2074, "completion_tokens": 45, 
"total_tokens": 2119}, "finish_reason": "tool_calls"}

<-- WatsonxChatModel[finish]:
{"messages": [{"role": "assistant", "content": [{"type": "tool-call",
"id": "chatcmpl-tool-2d9077c7a3564857845f856ab0484568", 
"tool_name": "retrieve_final_answer_from_memory", "args": 
"{\"memory_path\": 
\"extract_sds-3bec235a-8bb3-4e1f-b049-029c655f54f1/ingredients\"}"}
]}], "usage": {"prompt_tokens": 2351, "completion_tokens": 50, 
"total_tokens": 2401}, "finish_reason": "tool_calls"}

<-- WatsonxChatModel[finish]:
{"messages": [{"role": "assistant", "content": [{"type": "tool-call",
"id": "chatcmpl-tool-076d470dba0e42a0a860843d43b70478", 
"tool_name": "retrieve_final_answer_from_memory", "args": 
"{\"memory_path\": 
\"extract_sds-3bec235a-8bb3-4e1f-b049-029c655f54f1\"}"}]}], 
"usage": {"prompt_tokens": 2421, "completion_tokens": 48, 
"total_tokens": 2469}, "finish_reason": "tool_calls"}

<-- FunctionTool[retrieve_final_answer_from_memory][finish]: 
"{'product_name': 'Titanium(IV) oxide, anatase', 
'ingredients': [{'name': 'Titanium dioxide', 'cas_rn': '1317-70-0',
'concentration': '', 'formula': 'O2Ti'}]}"

The product contained in the PDF file is 'Titanium(IV) oxide, 
anatase'. The ingredients are:
- Name: Titanium dioxide
  - Chemical Formula: O2Ti
  - CAS Number: 1317-70-0
\end{verbatim}

\end{document}